\documentclass[letterpaper, 10 pt, conference]{ieeeconf}  

\IEEEoverridecommandlockouts               
\usepackage{cite}
\usepackage{graphicx}
\usepackage{makecell}
\usepackage{multirow}
\usepackage{amssymb}
\usepackage{hyperref}
\usepackage{amsmath,amsfonts} 

\usepackage{booktabs}

\usepackage{url}
\usepackage{float}
\usepackage{textcomp}

\usepackage{gensymb}
\usepackage{multicol,tabularx,capt-of}
\usepackage{multirow}

\title{\Large \textbf{3D-ADAM: A Dataset for 3D Anomaly Detection in Additive Manufacturing}}

\author{Paul McHard$^{1}$ $^{2}$, Florent P. Audonnet$^{1}$, Oliver Summerell$^{1}$, \\ Sebastian Andraos$^{2}$, Paul Henderson$^{1}$ and Gerardo Aragon-Camarasa$^{1}$
\thanks{$^{1}$ School of Computing Science, University of Glasgow, G12 8QQ, Scotland, United Kingdom {\tt\small p.mchard.1@research.gla.ac.uk; gerardo.aragoncamarasa@glasgow.ac.uk}}
\thanks{$^{2}$ HAL Robotics Ltd.,Unit 202, 115 Coventry Road, E2 6GG, England, United Kingdom {\tt\small p.mchard@hal-robotics.com, s.andraos@hal-robotics.com}}}

\begin{document}
\maketitle
\begin{abstract}

Surface defects are a primary source of yield loss in manufacturing, yet existing anomaly detection methods often fail in real-world deployment due to limited and unrepresentative datasets. To overcome this, we introduce 3D-ADAM, a 3D Anomaly Detection in Additive Manufacturing dataset, that is the first large-scale, industry-relevant dataset for RGB+3D surface defect detection in additive manufacturing. 3D-ADAM comprises 14,120 high-resolution scans of 217 unique parts, captured with four industrial depth sensors, and includes 27,346 annotated defects across 12 categories along with 27,346 annotations of machine element features in 16 classes. 3D-ADAM is captured in a real industrial environment and as such reflects real production conditions, including variations in part placement, sensor positioning, lighting, and partial occlusion. Benchmarking state-of-the-art models demonstrates that 3D-ADAM presents substantial challenges beyond existing datasets. Validation through expert labelling surveys with industry partners further confirms its industrial relevance. By providing this benchmark, 3D-ADAM establishes a foundation for advancing robust 3D anomaly detection capable of meeting manufacturing demands.

\end{abstract}

\section{Introduction}
The manufacturing sector constantly strives to reduce costs and improve quality and consistency through the use of advancements in technology.
A key goal is to automatically control and monitor the entire manufacturing life-cycle of a given part.
This includes detecting any anomalies or defects that may arise during the manufacturing process, and taking appropriate corrective action \cite{scheer2015}.

Accurately and reliably detecting surface defects during the manufacturing process remains a significant challenge across the manufacturing industry \cite{apqc2024}, affecting over 42\% of manufacturers. 2D Anomaly Detection methods are a mature and well established technology, with a diverse range of high-volume 2D Anomaly Detection Datasets available \cite{mvtec2021} \cite{SMAP2018} and a large body of work focused on tasks such as few shot anomaly detection \cite{WinCLIP2023} \cite{AnomalyCLIP2023} and unsupervised anomaly detection \cite{DFM2024} \cite{ContextFlow++2024}. In contrast, 3D Anomaly Detection methods lack the same degree of maturity. We attribute this to the fact that suitable datasets are limited and do not provide the breadth of features and defect examples required for developing solutions capable of performing at an acceptable level across the breadth of manufacturing contexts. Four leading datasets have been proposed previously for the challenge of 3D Anomaly Detection, MVTec3D-AD \cite{mvtec2022}, Eyecandies \cite{eyecandies2022},  Real3D-AD \cite{real3dad2023} and PAD \cite{pad2023}. Each of these datasets has several limitations which make them unsuitable for industrial applications, with the common issue between them being a lack of breadth in the defect and feature categories presented in the manufacturing samples.

\begin{figure*}[t]
 \centerline{\includegraphics[width=0.71\textwidth]{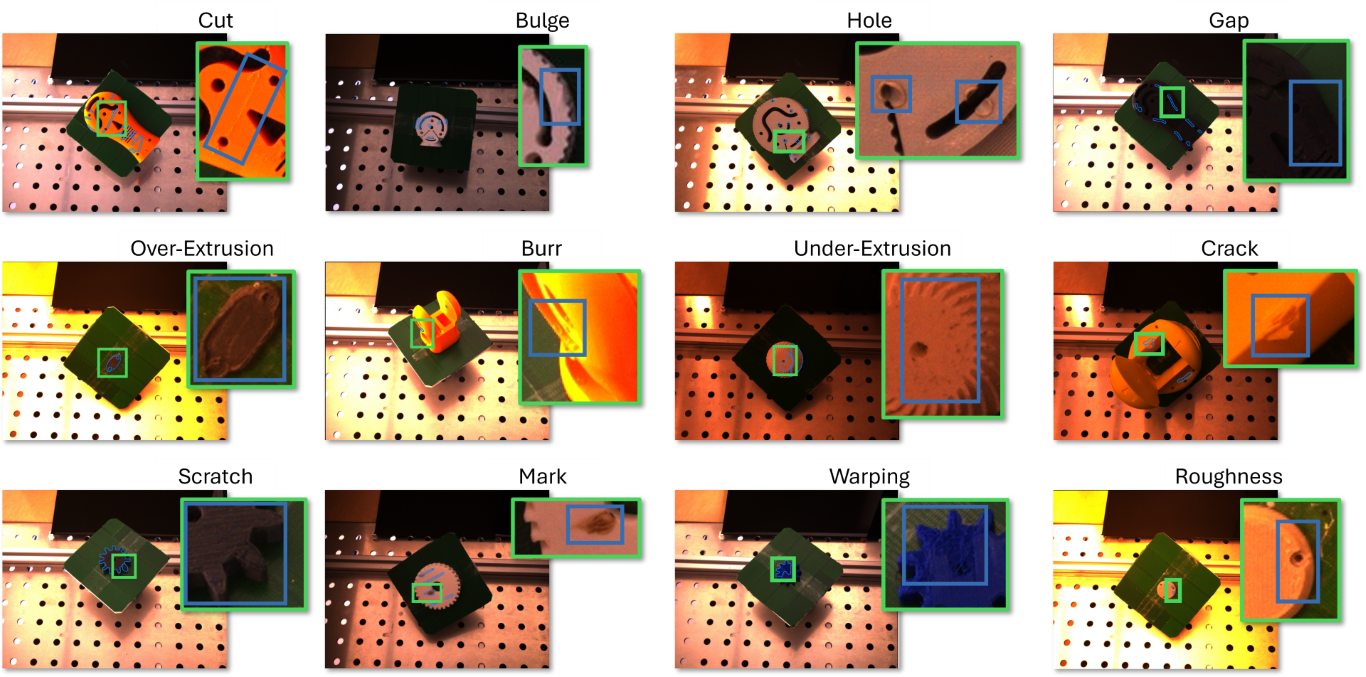}}
 \caption{Highlighted ground truth defect examples across each defect class.}
 \label{fig:defects}
 \end{figure*}
 
Hence, we introduce \textbf{3D-ADAM}, a large-scale, high-resolution, multi-camera dataset for 3D-Anomaly Detection in Additive Manufacturing. 3D-ADAM addresses the limitations of existing datasets by including comprehensive suite of examples of defects in 12 classes across a range of machine element features \cite{Mott2023}, which cover the range of surface defect types that may be encountered in additive manufacturing environments. 3D-ADAM proposes a set of 16 machine element features  \cite{Mott2023} across all part categories, with care taken in curating a range of defect examples displayed on each machine element type. 

The manufactured parts contained in 3D-ADAM constitute a complete assembly for a robot arm mechanism, based on the open-source BCN3D-Moveo system \cite{BCNMoveo2022} and augmented with additional components sourced from the Thingi10K \cite{Thingi10K2016} part dataset in order to ensure all relevant machine elements are adequately represented. We manufactured all the parts, both good and defective, in Polylactic Acid (PLA) as it is amongst the most commonly used materials for Fused Deposition Modelling (FDM) based additive manufacturing. Some defects naturally arose during the printing process, while others were forced by adjusting the printing parameters or by manual intervention. We assess the performance of our dataset in the tasks of 2D, 3D and RGB + 3D Anomaly Detection in an unsupervised setting across a suite of leading models, demonstrating that the 3D-ADAM dataset presents a novel challenge for current leading architectures in 3D and RGB+3D Anomaly Detection, which will bridge the gap between the current SOTA in 3D Anomaly Detection models and the level of performance required for deployment in real-world scenarios. Our main contributions are: 

\begin{itemize}
    \item We propose the first multi-camera, multi-modal, industry-relevant dataset for 3D and RGB+3D Anomaly Detection and Segmentation. Comprising 14,120 high-resolution image and 3D point cloud scans across 28 categories, with 27,346 annotated defect regions in 12 classes and a further 27,346 annotated machine element features in 16 classes.
    \item We evaluate foundational and leading methods in unsupervised 2D, 3D and RGB+3D Anomaly Detection and Segmentation tasks. Our initial benchmark demonstrates that these methods under-perform on our dataset, and that there is significant room for improvement for such methods to be suitable for real-world environments.
    \item We evaluate the quality of our ground-truth defect annotations through a survey of industry experts which shows that individuals with expert knowledge of surface defects in additive manufacturing contexts strongly agree with the image-level quality and pixel-level accuracy of the annotations proposed in our dataset.
\end{itemize}

\section{Related Work}

\begin{table*}
    \centering
    \caption{Specifications of sensors used to capture the 3D-ADAM dataset.}
\resizebox{0.8\textwidth}{!}{%
    \begin{tabular}{@{}lllll@{}}
    \toprule
        \textbf{Sensor} & \textbf{LSR-L} & \textbf{Nano} & \textbf{RealSense D455} & \textbf{Zed2i} \\
        \midrule
        \textbf{Manufacturer} & Mech-Mind & Mech-Mind & Intel & Stereolabs \\ 
        \textbf{Sensor Type} & Sony CMOS & Sony CMOS, & Intel SDM & Stereo Depth Sensor \\
        \textbf{Resolution (pix)} & 2048 x 1536 & 1280 x 1024 & 1280 x 720 & 1920 x 1080 \\
        \textbf{Optimal Range (mm)} & 1500 - 3000 & 300 - 600 & 600 - 6000 & 300 - 1200 \\ 
        \textbf{FOV (mm)} & 1500 x 1200 @ 1.5m & 220 x 150 @ 0.3m & 825 x 580 @ 0.6m & 490 x 345 @ 0.3m \\
        ~ & 3000 x2400 @ 3.0m & 440 × 300 @ 0.6 m & 8260 x 5820  @ 6m & 1965 x 1375 @ 1.2m \\
        \textbf{Precision (mm)} & 0.5-1.0 & 0.1 & 80 & 30 \\ 
        \bottomrule
    \end{tabular}}
    \label{tab:table1}
\end{table*}

\subsection{Additive Manufacturing Defects}

Additive Manufacturing encompasses a breadth of fundamentally different processes. In their extensive review of additive manufacturing processes and their effects on part defects, de Pastre et al. \cite{dePastre2022} categorise these into seven distinct processes; Binder Jetting, Directed Energy Deposition, Material Extrusion, Material Jetting, Powder Bed Fusion, Sheet Lamination and Vat Photo-Polymerisation. In the context of industrial robotics applications, it is only Material Extrusion processes, in particular Fused Deposition Modelling (FDM), which are relevant. Within the bounds of FDM processes, literature concludes that geometry is of more significant consideration than material when considering both the likelihood and form of defects occurring during the manufacturing process \cite{dePastre2022}, \cite{papazetis2018}.

\subsection{3D Anomaly Detection Datasets}
Four datasets have previously been proposed for 3D Anomaly Detection, MVTec3D-AD \cite{mvtec2022}, Eyecandies \cite{eyecandies2022}, Real3D-AD \cite{real3dad2023} and PAD \cite{pad2023}. Of these, MVTec 3D-AD and Real3D-AD represent the current leading datasets used as benchmarks for the problem of 3D Anomaly Detection in manufacturing. The Eyecandies dataset \cite{eyecandies2022} is comprised of synthetically generated parts and defects on object instances that are unsuitable for industrial contexts, and is thus not viable for such real-world applications. Similarly, the PAD dataset \cite{pad2023}, presents novel approaches to dataset acquisition for 3D Anomaly Detection, however the foundation of their dataset is assembled LEGO models, with defects relevant only to that context, with little extension into the broader manufacturing landscape. In the Real 3D-AD Dataset, Liu et al. \cite{real3dad2023} claim their dataset is more suitable for achieving high-precision 3D Anomaly Detection, supported by their notably higher point resolution and precision when compared to the MVTec 3D-AD dataset. However, the MVTec dataset remains the most popular among new techniques published for 3D Anomaly Detection. Both datasets feature a similar number of object categories, 10 in the MVTec 3D-AD Dataset and 11 in the Real3D-AD Dataset, respectively, with the volume of scan data per object category also comparable. The key distinction between the two is that the MVTec dataset presents a superior and more rigorous ground truth annotation methodology, and thus remains more widely adopted. Table \ref{tab:table_comp} compares the size and complexity between our dataset, MVTec3D-AD and Real3D-AD.

\renewcommand{\arraystretch}{1.6} 
\begin{table}[t]
    \centering
    \caption{Dataset Comparison between SOTA and 3D-ADAM}
    \resizebox{0.5\textwidth}{!}{
    \begin{tabular}{@{}c|*{4}{c}@{}}
    \hline
        \makecell{\textbf{Dataset}} & \textbf{MVTec3D-AD} & \textbf{Real3D-AD} & \textbf{3D-ADAM (Ours)} \\ 
    \hline
        \makecell{\textbf{\# of Total} \textbf{Scans}} & 4,147 & 1,254 & \textbf{14,120} \\ 
        \makecell{\textbf{\# of Categories}} & 10 & 12 & \textbf{28} \\
        \makecell{\textbf{\# of Defect} \textbf{Classes}} & 8 & 2 & \textbf{12} \\ 
         \makecell{\textbf{\# of Defect} \textbf{Annotations}} & 1,148 & 602 & \textbf{27,346} \\ 
        \makecell{\textbf{\# of Machine Element}\\\textbf{Classes}} & - & - & \textbf{16} \\ 
        \makecell{\textbf{\# of Machine Element}\\\textbf{Annotations}} & - & - & \textbf{27,346} \\ 
        \makecell{\textbf{Setting}} & \makecell{Single-Sensor, \\Multi-view} & \makecell{Single-Sensor, \\Multi-view} & \makecell{\textbf{Multi-Sensor,}\\\textbf{Multi-view}} \\ 
        \makecell{\textbf{\# Sensors}} & 1 & 1 & \textbf{4}\\ 
        \makecell{\textbf{Annotation Type}} & Single Mask & Single Mask & \textbf{Per-Defect Mask} \\ 
    \hline
    \end{tabular}
    }
    \label{tab:table_comp}
\end{table}

\subsection{Industrial Anomaly Detection}


The current state-of-the-art in 3D Anomaly Detection \cite{3dsr2024} \cite{transfusion2024} \cite{glfm2025} presents a number of highly capable models for unsupervised 3D anomaly detection. These models have all been trained, tested and validated on the MVTec 3D-AD Dataset \cite{mvtec2022}. Each of these models is underpinned by backbones built on the PointNet \cite{qi2017} and ResNet \cite{resnet2015} models, however vary significantly in the architectures proposed.

Zvartanik et al. \cite{3dsr2024} propose the Depth-Aware Discrete Auto-Encoder (DADA) architecture, which takes both the RGB and Depth image data jointly as a single input, enabling the model to learn in a general discrete latent space. The DADA architecture is the core of their 3D Dual Subspace Reprojection Network (3DSR) \cite{3dsr2024}, which is an extension of the authors' previous work \cite{zavrtanik2022}, in using dual-encoder models for 2D anomaly detection.  Conversely, a leading model in 3D Anomaly Detection proposed by Cao et al. \cite{cpmf2024} does not rely on raw RGB information as input in any capacity. Instead, they propose a Pseudo Multi-Modal Feature, in which 2D features are extracted by creating projections of the 3D point cloud data into 2D at a number of view-points, relying on pre-trained 2D image networks. These multi-view features are then aggregated together with hand-crafted 3D features to construct the memory banks from which inference can be performed for anomaly detection. Fucka et al. \cite{transfusion2024} propose a model that leverages modern diffusion-based techniques in a transparency-based diffusion model for 3D Anomaly Detection. Lastly, Cheng et al. \cite{glfm2025} have released the most recent and most promising model to date, which uses a PatchCore \cite{patchcore2021} based backbone to produce both global and local feature mappings, which their Adapted Point Transformer then utilises in conjunction with one another to perform Anomaly Detection and Segmentation predictions.

On the MVTec 3D-AD Dataset upon which the leading models are all trained, Cheng et al.'s GLFM model \cite{glfm2025}, currently has the hightest performing results on both Detection and Segmentation tasks in RGB+3D Anomaly Detection, whilst Fucka et al.'s \cite{transfusion2024} TransFusion model performs best on 3D Segmentation benchmarks, and Cao et al.'s \cite{3dsr2024} CPMF model provides the strongest performance on 3D Detection Benchmarks, amongst available models.

\section{The 3D-ADAM Dataset}

The 3D-ADAM dataset consists of 14,120 scans distributed equally between 4 high-resolution industrial 3D sensors. The sensors deployed are MechMind LSR-L, MechMind Nano, Stereolabs Zed 2i, and Intel Realsense (as shown in Figure \ref{fig:experimental_setup}). The details of these sensors are described in Table \ref{tab:table1}. To obtain a comprehensive coverage of the technologies deployed in industrial vision applications, these sensors vary in quality, cost, capture resolution, and depth-sensing technology. 

We aim to provide exhaustive coverage of the range of relevant defect and machine element types, such that we offer a diverse range of defect examples for industrial anomaly detection applications. To achieve this, 3D-ADAM contains scans of parts from 28 distinct objects. The majority of these parts represent the complete assembly for a robot arm mechanism, obtained from the open-source BCN3D-Moveo system \cite{BCNMoveo2022}. This set is augmented with additional components sourced from the open-source Thingi10k dataset \cite{Thingi10K2016} to provide a full suite of machine elements, covering common elements such as faces and edges, internal and external fillets, internal and external chamfers, holes, kerfs, tapers, indents, counterbores, countersinks as well as the most common gear types; spur gear teeth, rack gear teeth, spiral gear teeth and both clockwise and counter-clockwise helical gear teeth, for a total of 16 distinct machine element classes. Examples of these machine element types can be found in the accompanying video.

In addition, we curate a full suite of defect examples across this breadth of machine element examples, distributed across the range of part categories. The range of defect classes is designed to represent the full spectrum of distinct surface defect types which can be encountered in additive manufacturing environments \cite{dePastre2022}, \cite{papazetis2018} and includes (shown in Figure \ref{fig:defects}); cuts, bulges, holes, gaps, burrs, cracks, scratches, marks, warping, roughness and, specific to the additive manufacturing process employed in producing this dataset, over extrusion and under extrusion defects, for a total of 12 defect classes, which is the largest range provided by any dataset to date. Table \ref{tab:table2} describes the breakdown of total parts, total scans and total number of defect and machine element annotations, as well as the number of unique defect classes and machine element classes featured in each category of our dataset.

\renewcommand{\arraystretch}{1.25} 
\begin{table}
    \centering
    \caption{Statistical Breakdown. \# of defect and \# of machine element classes is given per per part category. \label{tab:table2}}
    \resizebox{0.49\textwidth}{!}{%
    \begin{tabular}{@{}l|*{6}{c}@{}}
    \hline
    \textbf{Category} &
      \textbf{Parts} &
      \textbf{Images} &
      \textbf{Defects} &
      \makecell{\textbf{Machine}\\\textbf{Elements}} &
      \makecell{\textbf{Defect}\\\textbf{Classes}} &
      \makecell{\textbf{Machine Elem.}\\\textbf{Classes}} \\ \hline
    \textbf{Robot Base Joint}               & 10  & 672   & 600   & 600  & 2  & 6   \\
    \textbf{Robot Base Body}                & 11  & 704   & 1215  & 1215  & 6  & 8   \\
    \textbf{Robot Base Motor Housing}       & 16  & 972   & 2550  & 2550  & 8  & 6   \\
    \textbf{Robot Shoulder Body}            & 6   & 384   & 718   & 718  & 4  & 6   \\
    \textbf{Robot Shoulder Joint (Left)}    & 7   & 452   & 1404  & 1404  & 5  & 6   \\
    \textbf{Robot Shoulder Joint (Right)}   & 7   & 448   & 1336  & 1336  & 4  & 6   \\
    \textbf{Robot Forearm Body}             & 5   & 448   & 2152  & 2152  & 6  & 4   \\
    \textbf{Robot Elbow Joint (Left)}       & 6   & 384   & 1660  & 1660  & 4  & 5   \\
    \textbf{Robot Elbow Joint (Right)}      & 9   & 580   & 2614  & 2614  & 6  & 8   \\
    \textbf{Robot Wrist Body}               & 8   & 652   & 2085  & 2085  & 6  & 7   \\
    \textbf{Robot Wrist Joint (Left)}       & 9   & 576   & 1920  & 1920  & 5  & 8   \\
    \textbf{Robot Wrist Joint (Right)}      & 8   & 512   & 939   & 939  & 6  & 7   \\
    \textbf{Robot Shoulder Fixture (Left)}  & 6   & 384   & 54    & 54   & 2  & 1   \\
    \textbf{Robot Shoulder Fixture (Right)} & 6   & 384   & 62    & 62   & 2  & 1   \\
    \textbf{Robot Forearm Fixture}          & 4   & 256   & 0     & 0    & 0  & 0   \\
    \textbf{Robot Wrist Fixture}            & 4   & 256   & 0     & 0    & 0  & 0   \\
    \textbf{Robot Base Cover}               & 10  & 608   & 991   & 991  & 6  & 4   \\
    \textbf{Robot Shoulder Cover}           & 12  & 688   & 820   & 820  & 5  & 5   \\
    \textbf{Robot Elbow Cover}              & 8   & 448   & 527   & 527  & 2  & 5   \\
    \textbf{Robot Wrist Cover}              & 8   & 480   & 162   & 162   & 2  & 4   \\
    \textbf{Base Clamp Body}                & 2   & 124   & 0     & 0    & 0  & 0   \\
    \textbf{Base Clamp Bolt}                & 2   & 132   & 0     & 0    & 0  & 0   \\
    \textbf{Robot Gripper}                  & 11  & 896   & 338   & 338  & 4  & 3   \\
    \textbf{Helical Gear (CW)}              & 6   & 384   & 434   & 434  & 2  & 3   \\
    \textbf{Helical Gear (CCW)}             & 9   & 548   & 731   & 731  & 4  & 3   \\
    \textbf{Rack Gear}                      & 8   & 516   & 1146  & 1146  & 7  & 4   \\
    \textbf{Spiral Gear}                    & 9   & 560   & 1314  & 1314  & 5  & 5   \\
    \textbf{Spur Gear}                      & 10  & 672   & 1574  & 1574  & 6  & 4   \\ \hline
    \textbf{Total}                          & \textbf{217} & \textbf{14120} & \textbf{27346} & \textbf{27346} & -- & -- \\ \hline
    \end{tabular}
    }
\end{table}

 \subsection{Structure \& Data Acquisition}
 
 We employed additive manufacturing via Fused Deposition Modelling (FDM) \cite{cano2021} to manufacture the parts included in our dataset. For this task, four distinct systems were employed: two Prusa i3 systems, an Ender 3 Pro, and an Ultimaker 2, and all parts were manufactured in Polylactic Acid (PLA). These systems were used to produce one complete set of "good" parts, manufactured at optimal settings. Many of these "good" examples were ultimately found to exhibit some minor degree of surface defects, discovered during the annotation stage. The set of parts for defective cases was distributed randomly and evenly amongst the four systems. Defect instances were created either by manipulating manufacturing settings on the systems, in order to create manufacturing process-based defects such as bulges, gaps, warping, over extrusion and under extrusion. Surface defect examples of a mechanical nature, in contrast to those generated by process error, were generated by replicating the handling and manipulation errors. These examples result in a real manufacturing environment, including the generation of cuts, holes, cracks, marks, and scratches. The majority of instances of burrs and roughness defects were as a result of naturally occurring defective features on parts.

 \begin{figure}[t]
 \centering \includegraphics[width=0.4\textwidth]{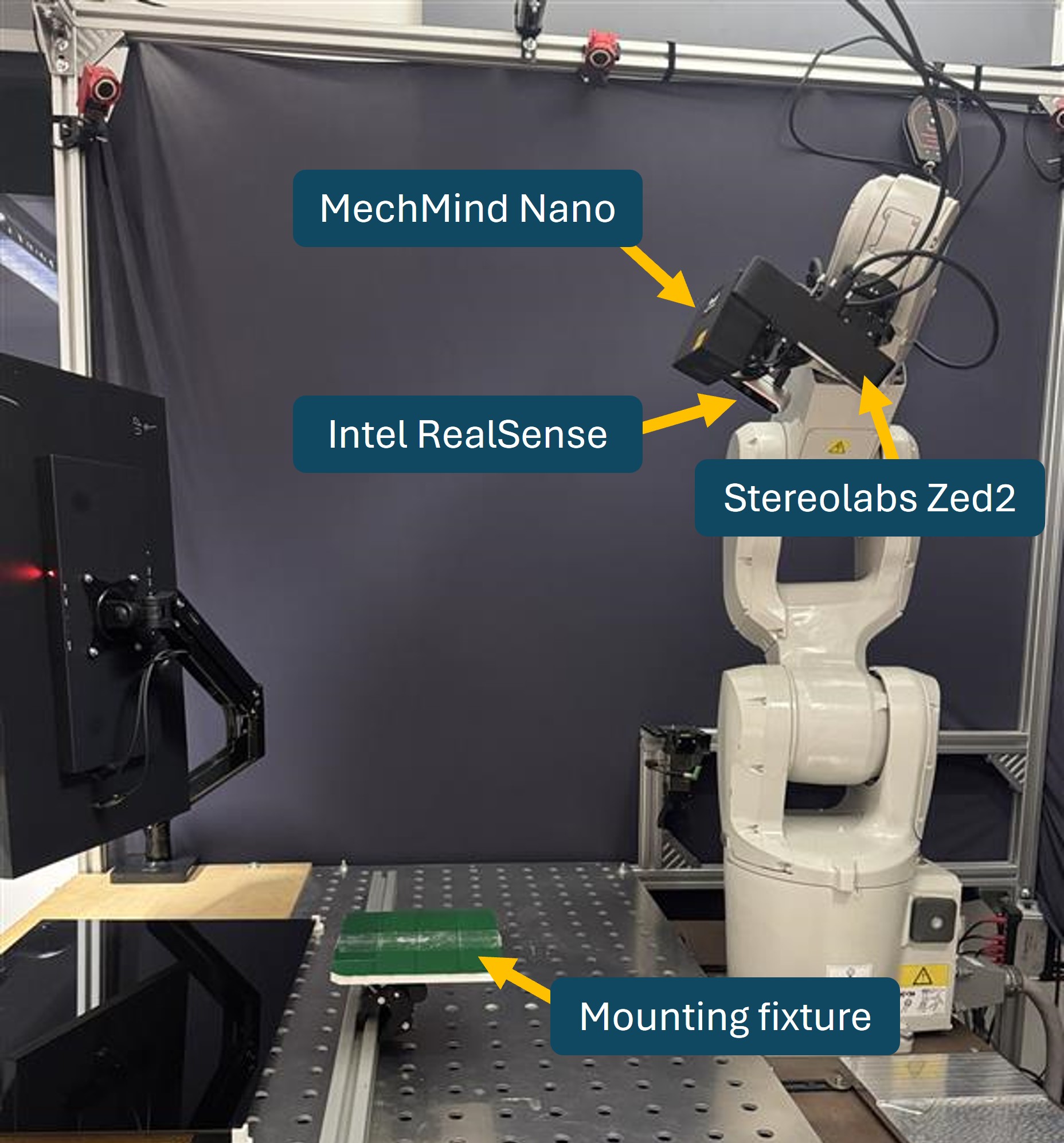}
 \caption{Experimental setup in laboratory conditions, showing mounting of the MechMind Nano, Intel RealSense and Stereolabs Zed2i sensors on an ABB IRB 120 robot arm, oriented towards the mounting fixture. Not pictured: MechMind LSR-L camera, mounted out of view above the cell frame.}
 \label{fig:experimental_setup}
 \end{figure}

 Data was acquired in a real-world environment in a robotics development cell. (Figure \ref{fig:experimental_setup}). Within this cell, an ABB IRB 120 robot arm was used to mount and orient the MechMind-Nano, Intel RealSense and Stereolabs Zed2i sensors, while MechMind LSR-L sensor was fixed to the upper exterior of the cell enclosure. The positions of all four industrial 3D imaging sensors were static throughout the data acquisition process. This mounting configuration was used to ensure that all four sensors would be operating within their optimal scanning ranges (as described in Table \ref{tab:table1}), and all four sensors were oriented towards a mounting fixture that held the parts in place during scanning. The mounting fixture allowed 2-axis rotation, with a 360-degree rotation around its primary axis and a 90-degree range of motion in its secondary axis. In each scanning procedure, the part was fixed at the centre of this fixture, and allowed to rotate through a range of orientations, with a minimum of a full rotation of the primary axis at 45-degree intervals, with the secondary axis fixed at both 0 and 30 degrees, respectively. This provided, in most instances, a complete view of the part. In those instances where additional coverage was desired, further variance in the secondary axis position was used to achieve the complete view. 

 \begin{figure*}[t]
\centering\includegraphics[width=0.8\textwidth]{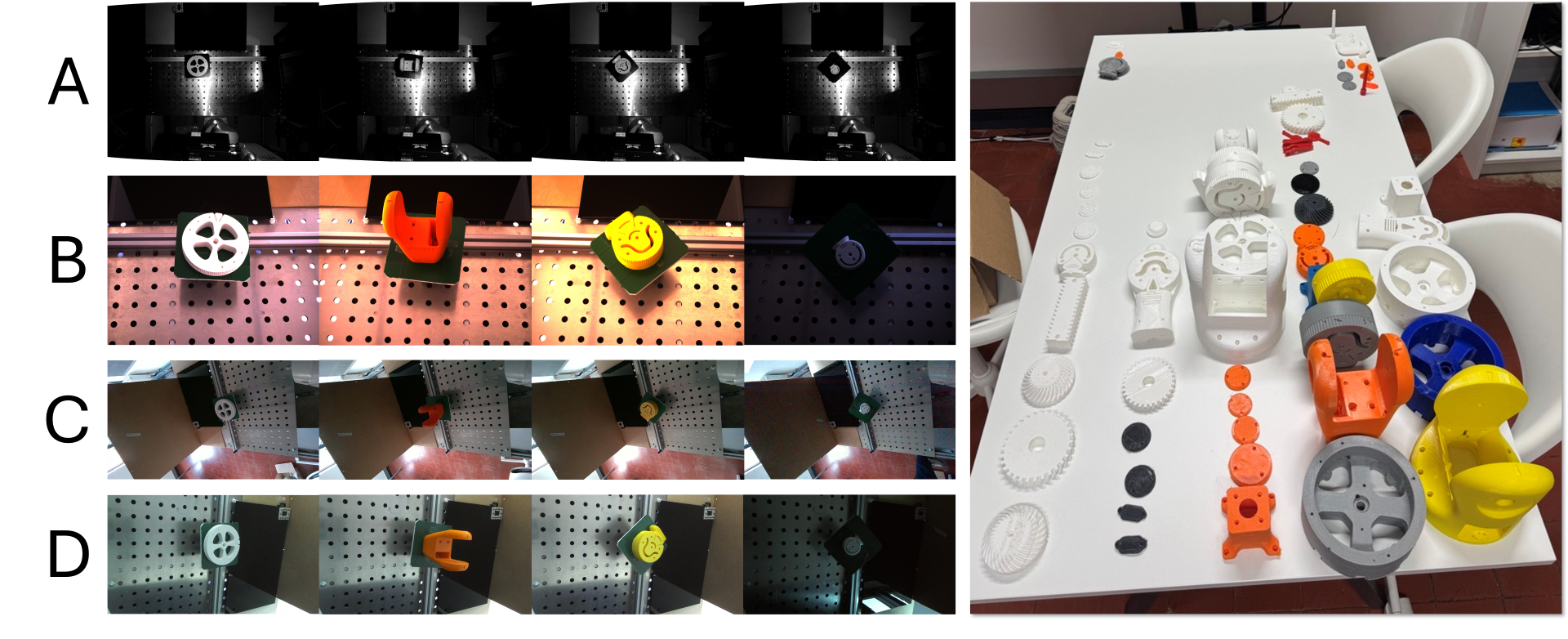}
\caption{Grid (right): Sample of images shown for selected parts in the subset shown, demonstrating differences in camera resolution, orientation and variations in lighting conditions. Images in row A were captured using the MechMind LSRL, while B, C and D, using the MechMind Nano, Intel RealSense and Stereolabs Zed 2i, respectively. Left: A sample subset of the dataset arrayed for capture on a table adjacent to the robot cell.}
\label{fig:sample_set}
 \end{figure*}

Each of the industrial scanners deployed captured six-channel images in the capture resolutions defined in Table \ref{tab:table1}, with these channels representing the \textit{r, g, b, x, y} and \textit{z} coordinates with respect to the sensors' local reference coordinate frame. These [\textit{x, y, z}] vectors represent the point cloud scan obtained by the sensor, and in all instances are captured as \texttt{.PLY} files. Additionally, for all sensors, the corresponding [\textit{r, g, b}] values for each pixel are captured as a \texttt{PNG} RGB image, which enables us to have a 1-to-1 mapping between the RGB and XYZ image data for each scan. The scene captured within our environment was open to indirect natural lighting sources during the period of scanning, as well as diffuse indirect artificial light sources when natural illumination levels were insufficient for proper data acquisition conditions. Thus, as scanning was conducted throughout the day, across weather and seasonal change variations, the data captured represents a wide range of natural and artificial lighting conditions as can be observed in Figure \ref{fig:defects} and \ref{fig:sample_set}.

\subsection{Ground Truth Annotations}
\begin{figure*}[t]
    \centering
    \includegraphics[width=0.85\textwidth]{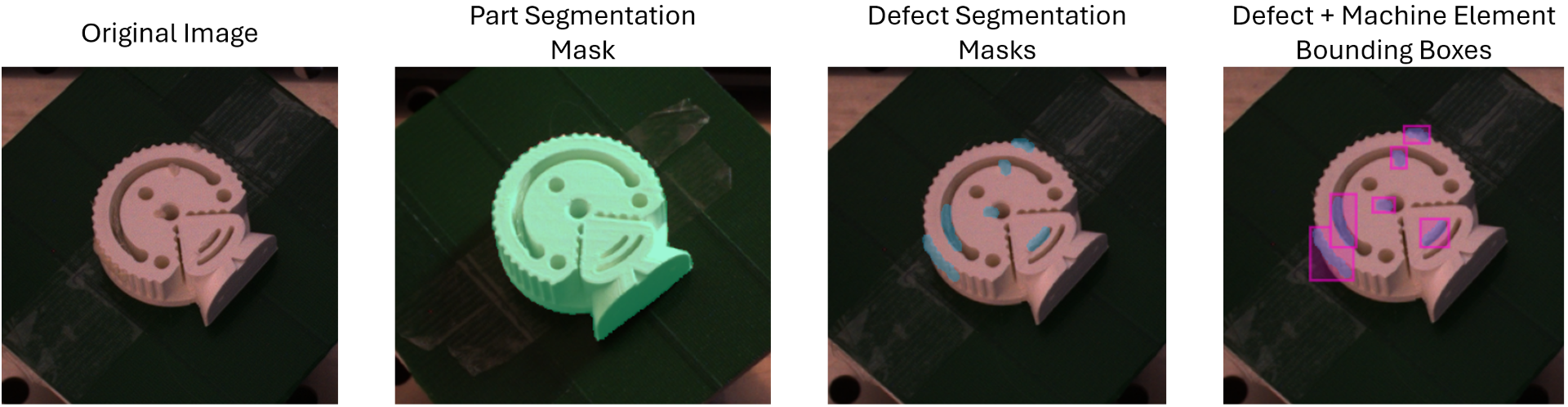}
    \caption{Visualization of Part Segmentation Mask, Defect Segmentation Masks and Machine Element Bounding Box annotations.}
    \label{fig:annotations}
\end{figure*}

We provide accurate, high-detail ground truth annotations for each instance of a part, machine element and defect in the dataset. Part annotations are provided as a segmentation mask, machine element annotations are a bounding box with a class label and defect annotations are a segmentation mask with a class label. These are demonstrated for an example part in Figure \ref{fig:annotations}.

For all cameras, a semi-automated segmentation method, based on Cutie \cite{cutie2023}, was employed to create foreground masks isolating the part in the scene. These were checked by hand and corrected where necessary. Defect and Machine Element ground truth annotations required expert knowledge of the manufacturing defects present in a part, and the machine elements on which they occurred. For this reason, all defect segmentation masks and machine element bounding boxes were annotated by hand on the 2D image data of each scan using the set captured from the MechMind Nano, as this sensor provided the highest level of detail of defects. A single expert annotator performed the annotations for all images in the MechMind Nano set, with verification and feedback support provided by domain experts. Homographic transform methods were employed to complete the ground truth annotations. These methods calculated the transformation matrix between part masks from the Mech-Mind Nano image to the equivalent from each of the remaining sensors, applying this transform to the defect masks and machine element bounding boxes to propagate the annotations from the MechMind Nano image set to the remaining systems with these being checked by hand with minor corrections performed where necessary.

\section{Benchmark and Evaluation}\label{sec:eval}

\subsection{Evaluation Methodology}
We evaluate our dataset on a suite of models in an unsupervised learning setting at the tasks of RGB, 3D and RGB+3D Anomaly Detection and Segmentation. An unsupervised training regime was selected as it presented the current strongest performing suite of state of the art models in the literature. For these tasks, we prepare our dataset in a suitable unsupervised training protocol, dividing into 3 sets; an anomaly-free training set, containing all of the defect-free scans captured for the dataset, with the annotated defective scans partitioned into a test and validation set in a 60:40 share. Partitioning is performed on a part-instance level such that view-point leakage is prevented within the protocol.

Using the \textit{anomalib} library \cite{anomalib2022}, a set of three foundational and leading available models for 2D Anomaly Detection were evaluated, namely PatchCore \cite{patchcore2021}, which we also evaluate in 3D protocol, UniNet \cite{uninet2025} and DinoMaly \cite{dinomaly2025}. To evaluate how existing 3D Anomaly Detection and Segmentation models perform on our dataset, as well as assess the challenge presented by the dataset, our benchmark includes the evaluation of three foundational models for 3D Anomaly Detection, as well as current state-of-the-art models for both 3D and RGB+3D Anomaly Detection and Segmentation. For the foundational models, we evaluate CFA \cite{cfa2022}, PaDiM \cite{padim2021} and PatchCore \cite{patchcore2021}, all of which were trained and tested using the \textit{anomalib} library \cite{anomalib2022}. Further, we include the three currently leading or recently leading available models at 3D and RGB+3D Anomaly Detection and Segmentation tasks on the MVTec3D-AD Dataset \cite{mvtec2022}. These are Transfusion \cite{transfusion2024}, 3DSR \cite{3dsr2024} and GLFM \cite{glfm2025}.

\subsection{Evaluation Metrics} 
To provide a strong foundation for benchmarking the performance of the models on our dataset at the tasks of RGB, 3D and RGB+3D Anomaly Detection and Segmentation, standard evaluation metrics for anomaly detection are employed. This maintains a consistent evaluation criteria across models, which allows accurate assessment of each model's performance on our dataset. These standard criteria are defined on the image-level anomaly detection performance and the pixel-level anomaly localisation performance. For detection and localisation tasks, the image-level and pixel-level Area Under Receiver Operator Curve (AUROC) metrics are utilised respectively. For models in which segmentation tasks were available, pixel-level anomaly segmentation was evaluated using the standard Area Under Per Region Overlap (AUPRO) metric. Further, for those models evaluated using the \textit{anomalib} library \cite{anomalib2022}, performance metrics in the form of both image and pixel level F1 scores are also provided.

\subsection{Results}
Table \ref{tab:table3} shows the results of the evaluation of 2D Anomaly Detection techniques on the 3D-ADAM dataset, noting strong performance at both image-level detection (Image AUROC) and pixel-level localisation (Pixel AUROC), consistent across models. Image F1 is considerably stronger in image-level detection tasks, and this is consistent across 2D techniques.

\begin{figure*}[t]
\centering\includegraphics[width=0.85\textwidth]{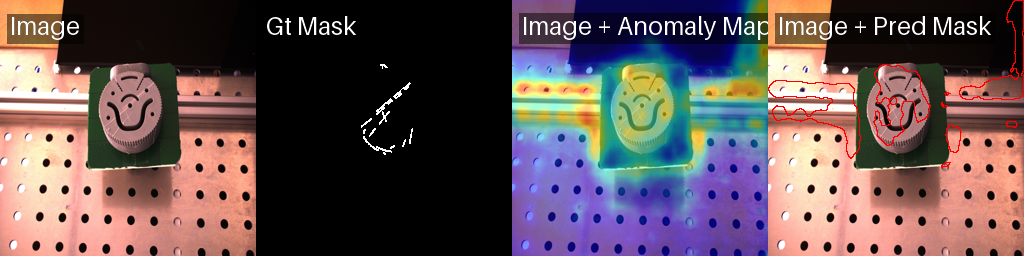}
\caption{Comparison of ground truth annotations with anomaly detection results achieved by PatchCore.}
\label{fig:model_visualisation}
\end{figure*}

Table \ref{tab:table4} shows the results of the evaluation of 3D and RGB+3D Anomaly Detection and Segmentation techniques on 3D-ADAM. It is notable here that foundational models perform well at detection (Image AUROC) and localisation (Pixel AUROC), with a similar trend of considerably stronger precision and recall in image-level detection (Image F1) as compared to pixel-level localisation (Pixel F1). Evaluation of the three leading models for 3D Anomaly Detection and Segmentation showed that models struggled significantly across available metrics as compared to their respective published results on the MVTec3D-AD Dataset, with the best performance being obtained by the GLFM model. Missing values, indicated by "----" in Tab. \ref{tab:table4} are attributable to the fact that the foundational models deployed through the \textit{anomalib} library \cite{anomalib2022} do not perform segmentation tasks, and therefore no AUPRO metric is calculated. Similarly, F1 scores are not computed by the three leading models; the authors of these respective models do not provide such metrics in their published evaluation, and we aim to take an unbiased approach for the assessment of these leading models. As such, F1 metrics for both Image and Pixel-level are respectively not computed. Figure \ref{fig:model_visualisation} shows a visualisation of the anomaly predictions obtained from the 3D implementation of PatchCore \cite{patchcore2021}, demonstrating that the model struggles significantly in accurately identifying defects as provided in the ground truth masks. Further examples of prediction results across the suite of models deployed in our benchmark can be found in the accompanying video.

We hold that the findings of this evaluation demonstrate that the 3D-ADAM dataset provides significantly greater challenge to leading models in the field compared to other SOTA datasets. We attribute this difference to the in-the-wild nature of our dataset, including the variations to ambient lighting conditions included in the model, as well as the superior, industry-standard set of defect annotations we provide. This evidence further supports the challenge posed by our dataset and the opportunity to train models suitable for real-world deployment. 

\begin{table}
    \centering
    \caption{Results for 2D Anomaly Detection and Localisation models on 3D-ADAM}
     \resizebox{0.47\textwidth}{!}{%
    \begin{tabular}{l|ccccc}
        \textbf{Model} & \textbf{Image AUROC} & \textbf{Image F1} & \textbf{Pixel AUROC} & \textbf{Pixel F1} \\ \hline
        \textbf{PatchCore} & 0.962 & 0.959 & 0.865 & 0.069 \\
        \textbf{UniNet} & 0.924 & 0.923 & 0.899 & 0.053 \\ 
        \textbf{DinoMaly} & 0.834 & 0.903 & 0.959 & 0.200  \\ 
        \textbf{Mean} & \textbf{0.907} & \textbf{0.929} & \textbf{0.908} & \textbf{0.107} \\
    \end{tabular}
    }
    \label{tab:table3}
\end{table}

\begin{table}
    \centering
    \caption{Results for 3D and RGB+3D models Anomaly Detection, Localisation and Segmentation tasks on 3D-ADAM}
     \resizebox{0.47\textwidth}{!}{%
    \begin{tabular}{l|ccccccc}
        \textbf{Model} & \textbf{Image AUROC} & \textbf{Image F1} & \textbf{Pixel AUROC} & \textbf{Pixel F1} & \textbf{AUPRO} \\ \hline
        \textbf{CFA} & 0.794 & 0.939 & 0.870 & 0.058 & ----- \\
        \textbf{PaDiM} & 0.964 & 0.975 & 0.846 & 0.041 & ----- \\ 
        \textbf{PatchCore} & 0.976 & 0.977 & 0.842 & 0.061 & ----- \\
        \hline \hline
        \textbf{3DSR} & 0.485 & ----- & 0.494 & -----  & 0.184 \\
        \textbf{TransFusion}  & 0.633  & ----- & 0.644 & ----- & 0.310 \\
        \textbf{GLFM}  & 0.791 & ----- & 0.820 & ----- & 0.603 \\
    \end{tabular}
    }
    \label{tab:table4}
\end{table}

\subsection{Industry Expert Survey} \label{sec:IndExSurv}

To validate the accuracy and suitability of our annotation methodology for industrial applications, we conducted a survey with industry experts, who were asked to re-annotate a subset of our data. Five industry experts from three different manufacturing organisations, across the industrial research and development, digital process manufacturing and robotics sectors from both private and public sector organisations, volunteered to take part in this survey. Each was provided with a randomised set of defective scans representing 1\% of the total dataset, with the intention that the task should be completable by a single expert in a few hours. A detailed set of instructions for the labelling task was provided, as well as "good" examples for the images assigned, providing as-designed, defect-free examples of each part instance alongside the defective case. Participants were not provided with any examples of annotated images from the core dataset, in order to minimise the degree of influence on their approach to annotations, and provide as close to a blind evaluation of the annotation process as possible. The ground truth annotations produced by experts were then evaluated for similarity to the equivalent ground truth annotations from the core dataset using a similar methodology as Section \ref{sec:eval}. However, given all labelling is conducted based on 2D image data, in place of the AUROC and AUPRO methods employed for Image-Level and Pixel-Level comparison respectively which is appropriate in 3D contexts, we instead use the Intersection over Union (IoU) criteria for image-level comparison and the PRO (Per Region Overlap) criteria.

The results of the labelling survey demonstrate strong correlation between the annotations within our dataset and those produced by industry expert surveys. Mean IoU score across all surveyed experts was 0.6 with a standard deviation of 0.04, indicating good image-level agreement on overall defect annotations for a given sample, with scores among labellers being tightly clustered. A mean PRO score of 0.76 with a standard deviation of 0.058 indicates strong pixel-level agreement on the shape and location of individual defects, with results again exhibiting low variability between labellers. The accompanying video 
provides a visual example where we have taken an instance where the task sets of two labellers overlapped for this single example. Alignment between labellers and the ground truth is strong, with labellers demonstrating strong correlation of true positive pixels, and tending to favour false positives over false negatives, thereby indicating that industry experts scrutinised a slightly more rigorous threshold for considering a given region to be defective in comparison to that proposed in our dataset.

\section{Discussion} \label{sec:Discussion}

We find that across the set of models tested with our dataset, the results indicate that the 3D-ADAM dataset provides a significantly greater challenge in comparison to the performance of the same models on the MVTec3D-AD dataset. 
The 3D-ADAM dataset presents additional new challenges in 3D Anomaly Detection, Localisation and Classification which are not currently facilitated by available datasets. Such challenges include \textit{few-shot and zero-shot multi-modal anomaly detection and localisation}, \textit{multi-modal anomaly classification}, and \textit{machine element feature classification}. Additionally, while our benchmark explores unsupervised 2D and 3D Anomaly Detection techniques exclusively, this is in order to both explore the performance of current leading models in literature, as well as maintain a consistent protocol across the benchmark. Moreover, with the 3D-ADAM dataset we make available protocols such that Supervised, Few-Shot and Zero-Shot Anomaly Detection techniques may all be utilised with our dataset, and presents an exciting opportunity for future work. This goes beyond what is provided by currently available leading datasets.

\subsection{Limitations}
We acknowledge that opportunities for improvement exist beyond what is proposed by our dataset. By composing our dataset exclusively of PLA-based additively manufactured parts, we present only one type of texture for models to identify surface defects upon. In many high-value advanced manufacturing applications, subtle differences in the presentation of certain defects may occur due to differences in material properties. However, as discussed, the literature consensus is that geometry is of far greater consequence to the occurrence of defects than material within the same additive manufacturing process. The degree this would impact applications in real-world use cases has not been explored and presents an interesting opportunity for future study.

While the range of surface defects accounted for in our dataset is the most comprehensive of any dataset for 3D Anomaly Detection to date, there remains an array of potential defects which can occur in advanced manufacturing use-cases, well understood by experts in the manufacturing and materials science fields, which are not provided in this dataset. This presents an opportunity for further augmentations to accommodate a broader range of industrial applications. Lastly, our dataset proides a multi-view scan of each part included, but does not provide full 3D views. We do not consider this a significant limitation, as we consider the availability of such complete scans to be unrealistic in real-world applications and instead anticipate that the multi-view, multi-camera in-the-wild approach proposed is significantly more appropriate for such contexts.

\section{Conclusion}

In this work, we present 3D-ADAM, a novel dataset for 3D Anomaly Detection, which aims to present a challenge closely aligned with the scenarios encountered in real-world industrial applications for Additive Manufacturing. 3D-ADAM is the largest dataset for this task to date, hosting the broadest range of defects of any dataset for 3D Anomaly Detection. 3D-ADAM uniquely provides machine element annotations for each defect, which no previous dataset has proposed, and is unique in presenting this challenge through a multi-view, multi-camera, in-the-wild dataset. 3D-ADAM provides a significantly more challenging dataset than any prior dataset, and is unique in having its defect annotation methodology verified through a survey of industry experts.  

Given the importance of identifying defects in manufacturing environments as a sector-wide challenge, the development of novel solutions for anomaly detection based on 3D image data is crucial for delivering a new generation of technologies for fully automating the detection, classification, localisation and ultimately correction of manufactured parts, in-process. The results of our benchmarking evaluation demonstrate that current leading models are not well equipped for this task due to the limitations of existing datasets for training, testing and validation. Our dataset provides a significant increase in the challenge presented, bringing it in line with realistic real-world use-cases that would be faced on deployment, thereby contributing to the broader ambition of developing mature 3D anomaly detection solutions in line with what is currently available in 2D anomaly detection.

{
\small
{}
}

\end{document}